\documentclass[11pt]{article}
\usepackage{acl2014}
\usepackage{times}
\usepackage{url}
\usepackage{latexsym}
\usepackage{graphicx}
\usepackage[table]{xcolor}
\usepackage{gb4e}
\usepackage{qtree}
\usepackage{adjustbox}
\usepackage{multibib}
\title{A weakly supervised sequence tagging and grammar induction approach to semantic frame slot filling}

\author{\begin{tabular}{cccccc}
\multicolumn{2}{c}{Janneke van de Loo^{1}} & \multicolumn{2}{c}{Guy De Pauw^{1,2}} & \multicolumn{2}{c}{Walter Daelemans^{1}}\\\\
\multicolumn{3}{c}{^{1}CLiPS - University of Antwerp} & \multicolumn{3}{c}{^{2}TEXTGAIN}\\
\multicolumn{3}{c}{Belgium} & \multicolumn{3}{c}{Belgium}\\
\multicolumn{3}{c}{{\small {\tt \{firstname.lastname\}@uantwerpen.be}}} & \multicolumn{3}{c}{{\small {\tt guy@textgain.com}}}\\
\end{tabular}}

\date{}

\begin{document}
\maketitle
\begin{abstract}

This paper describes continuing work on semantic frame slot filling for a command and control task using a weakly-supervised approach. We investigate the advantages of using retraining techniques that take the output of a hierarchical hidden markov model as input to two inductive approaches: (1) discriminative sequence labelers based on conditional random fields and memory-based learning and (2) probabilistic context-free grammar induction. Experimental results show that this setup can significantly improve F-scores without the need for additional information sources. Furthermore, qualitative analysis shows that the weakly supervised technique is able to automatically induce an easily interpretable and syntactically appropriate grammar for the domain and task at hand.

\end{abstract}

\section{Introduction}

The basic task of a command and control (C\&C) interface is to convert a language user's unstructured input into some structured format that can be unambiguously interpreted by a controller. Except for the simplest of devices, this requires at some point in the processing chain some kind of syntactic component that is aware of such issues as compositionality and word order. For example, in the home automation C\&C phrase {\em turn on the lights}, the system needs to at least be aware that the input language is SVO to trigger the difference between action and patient. In many cases, this is done through the use of a context-free grammar in the decoding process \cite{Jurafsky95usinga}, as illustrated in Figure \ref{cfgex} for the task of controlling a television using natural language.

\begin{figure*}
\small
\begin{center}
\begin{tabular}{ll}
{\tt $<$sentence$>$} & {\tt = $<$volume\_command$>$ $|$ $<$channel\_command$>$}\\
{\tt $<$volume\_command$>$} & {\tt = (set $|$ change) volume [VOL] to $<$number$>$}\\
{\tt $<$channel\_command$>$} & {\tt = (select $|$ change to) channel [CH] ($<$number$>$ | $<$name$>$)}\\
{\tt $<$number$>$} & {\tt = one [1] $|$ two [2] $|$ three [3] $|$ four [4] $|$ five [5]}\\
{\tt $<$name$>$} & {\tt = BBC [4] $|$ CNN [2] $|$ EuroSports [1]}\\
\end{tabular}
\end{center}
\caption{Context-free grammar for a television command \& control interface.}
\label{cfgex}
\end{figure*}

While this approach has proved to be quite effective for many applications, such devices are limited in the sense that they require the user to adhere to the grammar and lexicon as predefined by the designers of the device. This paper describes research conducted in the context of the ALADIN project, which investigates a command and control interface that is not bound by predefined linguistic constraints, but rather adapts to the user in a language and domain independent way. The aim of the project is to find techniques that minimize the training phase for such a self-learning system, so that it can be easily deployed as a home automation C\&C system for people with a physical impairment, who are often challenged by (progressively aggravating) speech impediments. A self-learning C\&C interface that is able to automatically learn and adapt to an individual user's linguistic characteristics, can provide an important means for a physically impaired person to regain some independence in a domestic setting.
% * <walter.daelemans@uantwerpen.be> 2015-08-24T14:19:27.837Z:
%
%  Referentie of webadres?
%

In this paper, we will investigate the syntactic component of the ALADIN-system, where we define ``syntax'' loosely as a means to interface between the surface utterance and its underlying semantic representation. Our semantic representation of choice is that of a semantic frame with slots that need to be filled with values (Section \ref{data}). Previous research efforts show that this can be done with a fair degree of accuracy using a hierarchical hidden markov model architecture (Section \ref{related}) under a minimal amount of supervision. The research described in this paper improves on this accuracy, through means of retraining using two different techniques: a shallow approach using sequence labeling (Section \ref{seqlabel}) and a more traditional parsing approach employing stochastic context-free grammar induction (Section \ref{cfg}). We will quantitatively and qualitatively evaluate these techniques on the basis of comparative experiments (Section \ref{exp}) and finish with a discussion of the results and pointers to future work (Section \ref{conclusion}).

\section{The {\sc patcor} Corpus}
\label{data}

The experiments described in this paper were conducted on the {\sc patcor} corpus\footnote{\url{https://github.com/clips/patcor}}. {\sc patcor} is a collection of command and control utterances for the single-player card game {\em Patience}. This control task was chosen because the disambiguation of these types of sentences requires awareness of compositionality and word order, while at the same time the domain and vocabulary are fairly restricted. The corpus contains audio recordings of nine players, totaling around 3000 spoken commands (Belgian Dutch) and their associated action in the game (represented as a semantic frame; cf. infra). All of the spoken commands were manually transcribed. 

{\sc patcor} consists of a core data set of eight speakers, for which around 250 utterances were collected. We will refer to this core set as the {\sc p1-8} set. Additional data was gathered for one player to study the effect of increasing training sizes. This data set, referred to as the {\sc p9} set contains over 1,000 utterances. In the experiments below, {\sc p1-8} is used for development purposes, while {\sc p9} is used for evaluation in learning curve experiments. More details about the data set and the command structures that were used by the speakers are described in \newcite{nlp4ita}. 

\begin{figure}
\begin{center}
{\em Put the jack of clubs on the red queen}
\adjustbox{valign=m}{ 
\includegraphics[width=4.5cm]{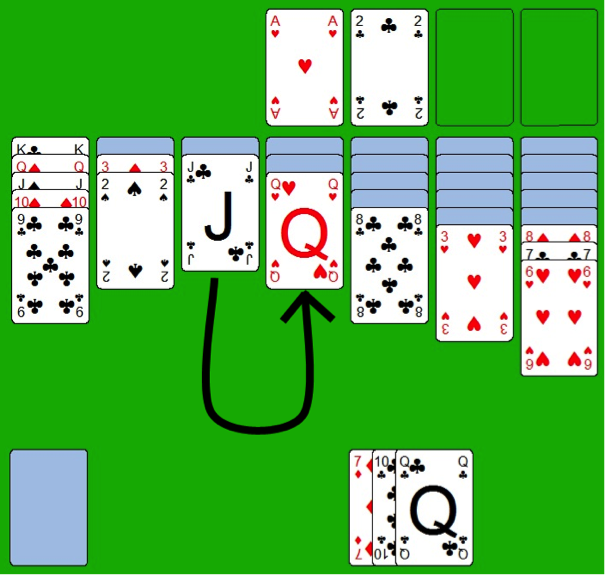}}
\small
\begin{tabular}{|l|l|}\hline
\multicolumn{2}{|c|}{{\em movecard}}\\\hline
Slot &	value \\\hline
$<$FS$>$&	c \\\hline
$<$FV$>$	& 11 \\\hline
$<$FF$>$&	-\\\hline
$<$FC$>$&	3 \\\hline
$<$FH$>$&	-\\\hline
$<$TS$>$&	h \\\hline
$<$TV$>$&	12 \\\hline
$<$TF$>$&	-\\\hline
$<$TC$>$&	4 \\\hline
\end{tabular}
\end{center}
\caption{Example of a Patience command in {\sc patcor}, the corresponding action on the playing field and the associated semantic frame. A move is defined as a combination of a from slot ($<$F?$>$) and a to slot ($<$T?$>$). A card is defined as the combination of a suit ($<$?S$>$) and a value ($<$?V$>$). Also defined are slots for the `hand' at the bottom ($<$FH$>$), the seven columns ($<$?C$>$) in the playing field and the four foundation stacks at the top right ($<$?F$>$).}

\label{patexample}
\end{figure}

The semantic representation of the controls in {\sc patcor} takes the form of semantic frames. This is illustrated in Figure \ref{patexample}: the semantic frame describes the card that is moved (as a suit/value pair), the card that it is moved onto, as well as the respective columns that are involved in the move. Additional slots are available for moves from the hand and the foundation (respectively at the bottom and top of Figure \ref{patexample}). An action on the playing field is thus described as a collection of slot-value pairs. There are two types of semantic frames: the {\em movecard} frame exemplified in Figure \ref{patexample} and an additional {\em dealcard} frame that denotes the action of asking for a new hand of cards. The latter frame has no slots or values to be filled.

Note that the automatically generated semantic frame is overspecified with respect to the command: in the command in Figure \ref{patexample}, columns are not mentioned, although this information is included in the automatically generated frame. We will use the utterances and their associated frames as training material to trigger the mapping from words onto slot-value pairs. The difficulty for the induction task therefore lies in identifying which words can be linked to which slots and associated values. Furthermore, a word might underspecify its slot-value in the semantic frame: for instance in Figure \ref{patexample}, the word ``rood'' ({\em red}) can refer to suits {\em hearts} and {\em diamonds} alike, while only the former is represented in the associated semantic frame.

\section{Related Research}
\label{related}

In this section we will provide some pointers to related research on mapping tokens to semantic frames in C\&C interfaces. We will also discuss previous research efforts on the {\sc patcor} data using two different techniques.

\subsection{Semantic frame induction}

There are many studies in spoken language understanding that use semantic frames or an equivalent to encode the meaning of utterances. Mapping new utterances to these semantic representations has been attempted using a variety of techniques, such as generatively associating word sequences with concept sequences \cite{Pieraccini1991,Wang2011}, using a discriminative technique to perform concept tagging \cite{Hahn2011}, or applying alignment techniques from the field of statistical machine translation to relate words with semantic concepts \cite{Epstein1996,Macherey2001}. 

The research described in this paper differs from these previous approaches in that our semantic representation is overspecified with respect to the utterance and that we do not assume the ready availability of gold-standard alignments between tokens and concepts. With respect to the latter issue, our research is more akin to the unsupervised alignment of utterances and concepts in the {\sc atis} corpus, as described in \newcite{Huet2011}. The {\sc patcor} task itself on the other hand, is more similar to that of the ROBOCUP data \cite{Chen2008}, for which a variety of approaches have been coined, ranging from learning the alignment between utterances and frames \cite{Liang2009} to learning semantic parsers \cite{Chen2010}. 

\subsection{Supervised concept tagging}
\label{nlp4ita}

Previous work on {\sc patcor} data is described in \newcite{nlp4ita} and \newcite{SMIAE}. These papers approach the problem of inducing semantic frame slots from (transcribed) utterances as a concept/sequence labeling task that can be learned by a machine learning classifier. To this end, the transcriptions of {\sc patcor} were manually annotated with IOB-concept tags, where each word is associated with a concept tag that denotes its association to a specific type of frame slot and its relative position in the text chunk ({\bf I}nside, {\bf O}utside or at the {\bf B}eginning). This is illustrated in the following command (English: {\em put the hearts five on the clubs six}) which denotes a move from a $<${\bf F}rom\_{\bf S}uit$>$ $<${\bf F}rom\_{\bf V}alue$>$ card to a $<${\bf T}arget\_{\bf S}uit$>$ $<${\bf T}arget\_{\bf V}alue$>$ card.

\begin{exe}
\ex leg$^{O}$ de$^{O}$ harten$^{I\_FS}$ vijf$^{I\_FV}$ op$^{O}$ de$^{O}$ klaveren$^{I\_TS}$ zes$^{I\_TS}$
\label{taggedexe}
\end{exe}

The eight data sets of the {\sc p1-8} set were each used individually as training data for a memory-based concept tagger and learning curves were plotted using the last 50 utterances in the respective data sets. Almost all of the data sets achieve an I-chunk F-score between 95\% and 100\% after around 130 training utterances, underlining the learnability of the task. However, this approach relies on the existence of a gold-standard mapping between the observed tokens and their relevant frame slots and values. This type of supervision is not available in a C\&C interface that is to be trained through the association of an utterance in its entirety with an (overspecified) semantic frame. Furthermore, \cite{nlp4ita,SMIAE} only report tag and chunk accuracy scores, which only gives us an indirect indication of how well semantic frame slots and values can be filled by this technique. In the experiments below, we consider this supervised sequence labeler as the upperbound of the level of performance that can be expected for this data set.
% * <walter.daelemans@uantwerpen.be> 2015-08-24T14:42:21.277Z:
%
%  Nog niet uitgelegd
%

\subsection{FramEngine}

Research on the acoustic part of {\sc patcor} focuses on finding descriptive word units in the utterances, using non-negative matrix factorization. To map observed units to frame slots (frame decoding), early research \cite{Gemmeke2012b} employed a hand-crafted grammar, which encoded expected word order patterns. This approach obtained slot-value recall scores between 37\% and 85\% on the {\sc p1-8} set. One of the major bottlenecks in achieving higher accuracies was the method's limited ability to build a robust acoustic representation on the basis of a limited amount of training data. 

To streamline frame decoding, a semantic frame induction engine was developed (\textit{FramEngine}) that uses a combination of non-negative matrix factorization and a hierarchical hidden markov model (\textit{HHMM}) to simultaneously detect recurring patterns in commands (words, morphemes, acoustic units) and relate them to their relevant slots and values in semantic frames. This is a weakly supervised induction task, as there is only supervision at the utterance level and no relations between the components of the utterances and the semantic frame slots are specified in advance. 

Previous work \cite{ons2013self} demonstrated the performance of an early implementation of this system on pathological speech input for a simple home automation command \& control task and non-pathological {\sc patcor} speech data. The results show that the system has a promising learning potential even with small amounts of training data, but that enhancements are needed in order to produce practically usable accuracies for more complex utterances, such as the ones found in {\sc patcor}. Work described in \newcite{1901.10680} focused on improvements with respect to the system architecture. Extensions regarding the HMM structure and a technique called ``expression sharing'' were added to FramEngine's workings and were shown to significantly improve on the frame-slot filling abilities on transcribed {\sc patcor} data.

\newcite{1901.10680} describes parameter selection experiments on {\sc p1-8}, while a final evaluation was conducted on the basis of learning curve experiments on the {\sc p9} set. FramEngine was able to score over 95\% frame-slot F-score using around 200 utterances as training data. While these results are encouraging, there are reasons to believe we can still increase accuracy by means of retraining. FramEngine's slot-value allocation technique is limited in the context that it can consider during disambiguation. And while word order is indirectly modeled through the structural setup of the hierarchical hidden markov model, a further explicitation of the syntactic properties of the observed utterances may provide additional insights into the induction process proper.

In the context of this paper, we refer to retraining as the process of using an automatically labeled data set as input data for another inductive processing step. Typically, retraining techniques attempt to label large volumes of unlabeled data to increase the volume of available training data ({\em self-training} \cite{mcclosky-charniak-johnson:2006:HLT-NAACL06-Main}) or adapt a classifier to a new domain ({\em co-training} \cite{Blum:1998:CLU:279943.279962}). The work in this paper investigates retraining as the process of using successive classifiers on the same training data to improve results. This approach has been successfully applied to a variety of problems, including morphological analysis \cite{DePauw2008d} and spoken language understanding \cite{wu-EtAl:2006:EMNLP}.

\section{Retraining}
\label{retraining}

This paper explores two different retraining techniques: inductive sequence labeling (Section \ref{seqlabel}) and stochastic context-free grammar induction (Section \ref{cfg}). 
As described in \newcite{1901.10680}, FramEngine can be shown to exhibit a more than decent performance on the task of frame-slot filling for the {\sc patcor} data. For retraining purposes, we can use FramEngine to decode its own training set, linking each word in the training set with a frame slot/value label through Viterbi decoding of the utterance. This means we can now also associate each word with an automatically assigned IOB-tag, expressing its frame-slot and in addition also its value:

\begin{exe}
\ex leg$^{O}$ de$^{O}$ harten$^{I\_FS=h}$ vijf$^{I\_FV=5}$ op$^{O}$ de$^{O}$ klaveren$^{I\_TS=c}$ zes$^{I\_TS=6}$
\label{autotagex}
\end{exe}

\noindent This information is consequently used as source data for retraining purposes.

\subsection{Sequence Labeling}
\label{seqlabel}

The first type of retraining technique recalls the research described in Section \ref{nlp4ita}: words annotated with IOB-tags are considered as training material for a discriminative sequence tagger. Fully supervised (i.e. with manually attributed IOB-tags), this approach can be observed to obtain near perfect accuracy scores on the {\sc patcor} data. Using training data labeled by a weakly supervised technique (i.e. FramEngine), a sequence labeler can be expected to perform less accurately than the fully supervised approach, but more accurately than FramEngine itself, since it can take into account more contextual information. We consider two different machine learning techniques driving the sequence labeler: we compare a memory-based tagger ({\sc mbt} \cite{mbt}) with one based on conditional random fields ({\sc wapiti} \cite{lavergne2010practical}). 

\begin{figure}
\includegraphics[trim=50 180 50 250,clip,width=\linewidth]{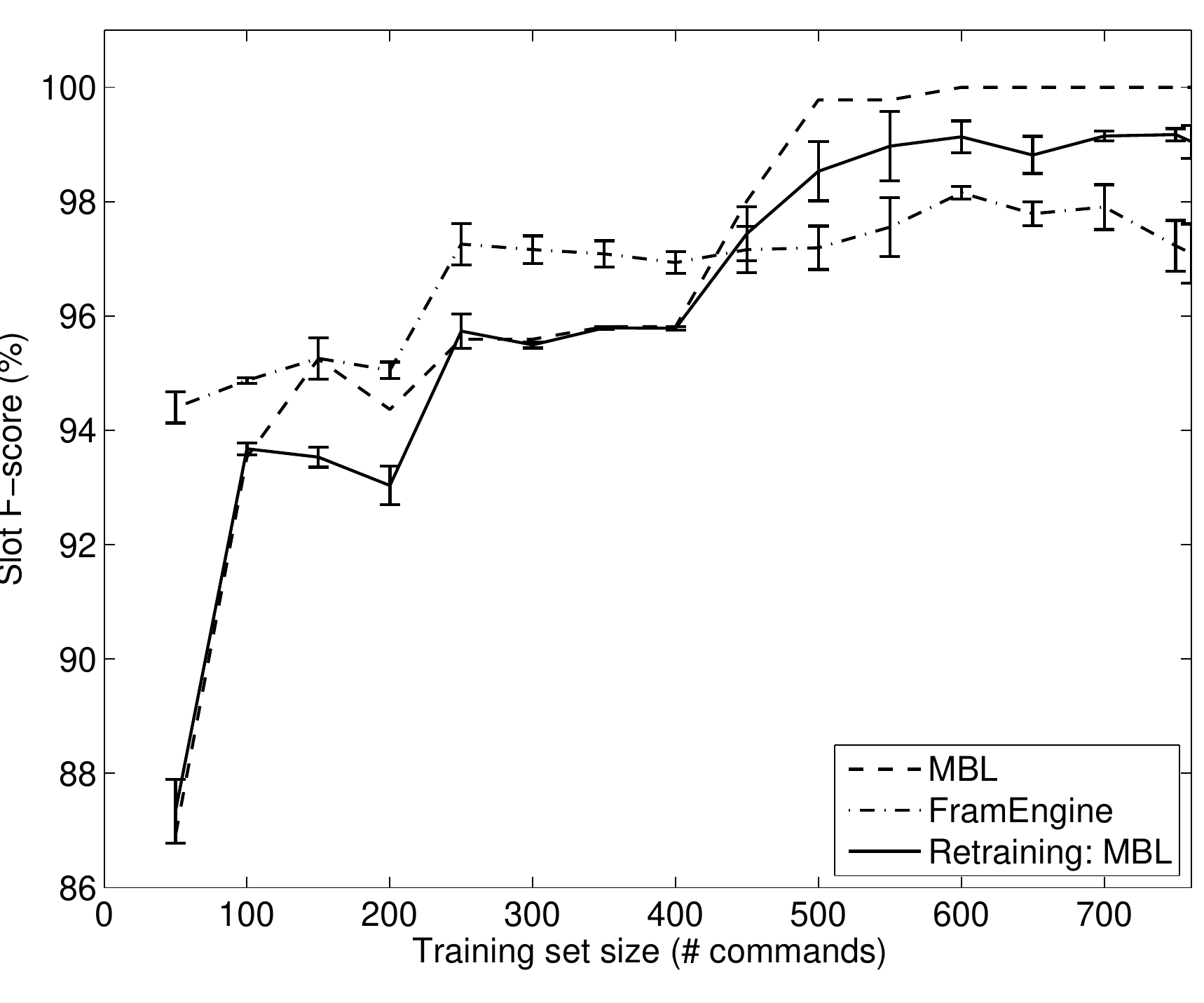}
\caption{Learning curves for {\sc p9} data FramEngine vs Single-Step Sequence labeling.}
\label{1step}
\end{figure}

Figure \ref{1step} displays learning curves for exploratory experiments, in which we compared fully supervised sequence labeling (using gold-standard labels), weakly supervised decoding using FramEngine, and retrained sequence labelers. As expected, the accuracy of the retrained sequence labeler eventually ends up somewhere between that of the supervised labeler and that of FramEngine. But this result is not obtained until rather late in the learning curve, and even the fully supervised labeler trails compared to the weakly supervised FramEngine until training data hits around 400 utterances. 

This is due to the fact that the sequence labelers consider the IOB-tags holistically. FramEngine as described in \newcite{1901.10680}, uses the method of {\em expression sharing}, in which the hidden markov model is aware of the fact that certain frame slots (such as the suit or value of a card) may be expressed using the same tokens, regardless of whether it occurs in a {\em from}-clause or a {\em target}-clause. This means that if a certain value has only occurred in one particular context in the training set (e.g. vijf$^{I\_FV=5}$), the tagger would not be able to predict the correct tag when it occurs in the other context in held-out data (e.g. vijf$^{I\_TV=5}$), as it is simply unaware this tag even exists. 

\begin{figure*}
\begin{center}
\includegraphics*[width=13cm]{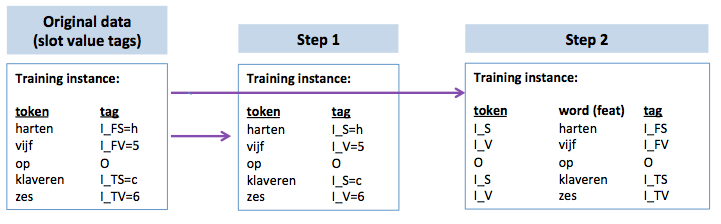}
\end{center}
\caption{Illustration of the two-step tagging approach.}
\label{aladinFramework}
\end{figure*}

We therefore implement a two-step tagging approach that implicitly introduces expression sharing in the workings of the sequence labelers. The process is illustrated in Figure \ref{aladinFramework}. The tags are decomposed into two parts: first a tagger attributes to the words slot-value tags that generalize over slots of the same type (in this case: ignoring `from' and `target' information). In the second step, a tagger is trained using the slot-portion of the tag from step 1 as the token to be tagged and the remainder of the original tag as the target tag. The original word in the utterance is used as an extra information layer towards disambiguation. The tags of both steps are then combined into the conglomerate tag. In case of incompatible tag-parts, the word is attributed the {\bf O}-tag. We will use the two-step tagging approach in the experiments described in Section \ref{exp}.

\subsection{PCFG Induction}
\label{cfg}

\begin{figure}
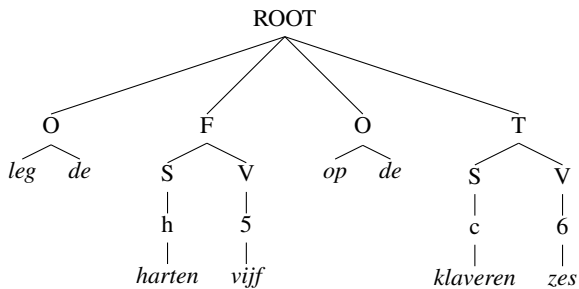

\small
%\Tree [.S [.O This ] [.VP [.V is ] [.NP a simple tree ] ] ]
\Tree [.ROOT [.O {\em leg} {\em de} ]  [.F [.S [.h {\em harten} ] ] [.V [.5 {\em vijf} ] ] ] [.O {\em op} {\em de} ] [.T [.S [.c {\em klaveren} ] ] [.V [.6 {\em zes} ] ] ] ] 
\caption{Phrase-Structure tree derived from Example \ref{autotagex}.}
\label{pstree}
\end{figure}

The utterances tagged automatically by FramEngine (cf. Example \ref{autotagex}) can also be transformed into tree-structures, given information about which expressions can be shared. Slot-values ({\tt =?} in Example \ref{autotagex}) can be considered as terminals, while suit/value and from/target indicators can function as nonterminals. Two consecutive clauses headed by the same nonterminal are joined. As such, the tagged sentence in Example \ref{autotagex} can be algorithmically transformed into the phrase structure in Figure \ref{pstree}. These phrase structures can then be transformed into a (probabilistic) context free grammar (Figure \ref{excfg}), by observing the productions used to construct the tree structures for the utterances in the training set. Finally, a Viterbi parser is used to assign tree structures to new utterances and a reversal of the tree construction process outlined above, can be used to assign IOB-labels that refer to frame slot/value pairs.

\section{Experimental Results}
\label{exp}

\begin{figure}
\small
\begin{center}
\begin{tabular}{lll}
{\tt ROOT} & $\rightarrow$ & {\tt O F O T}\\
{\tt F} & $\rightarrow$ & {\tt S V}\\
{\tt T} & $\rightarrow$ & {\tt S V}\\
{\tt S} & $\rightarrow$ & {\tt h $\vert$ c}\\
{\tt V} & $\rightarrow$ & {\tt 5 $\vert$ 6}\\ 
{\tt O} & $\rightarrow$ & {\em leg de $\vert$ op de}\\ \hline
{\tt h} & $\rightarrow$ & {\em harten}\\
{\tt c} & $\rightarrow$ & {\em klaveren}\\
{\tt 5} & $\rightarrow$ & {\em vijf}\\
{\tt 6} & $\rightarrow$ & {\em zes}\\
\end{tabular}
\end{center}
\label{excfg}
\caption{Context-free grammar sample and lexicon derived from Figure \ref{pstree}.}
\end{figure}

\begin{figure*}[ht]
\begin{center}
\begin{tabular}{cc} % L B R T
\includegraphics[trim=50 180 50 200,clip,width=7.5cm]{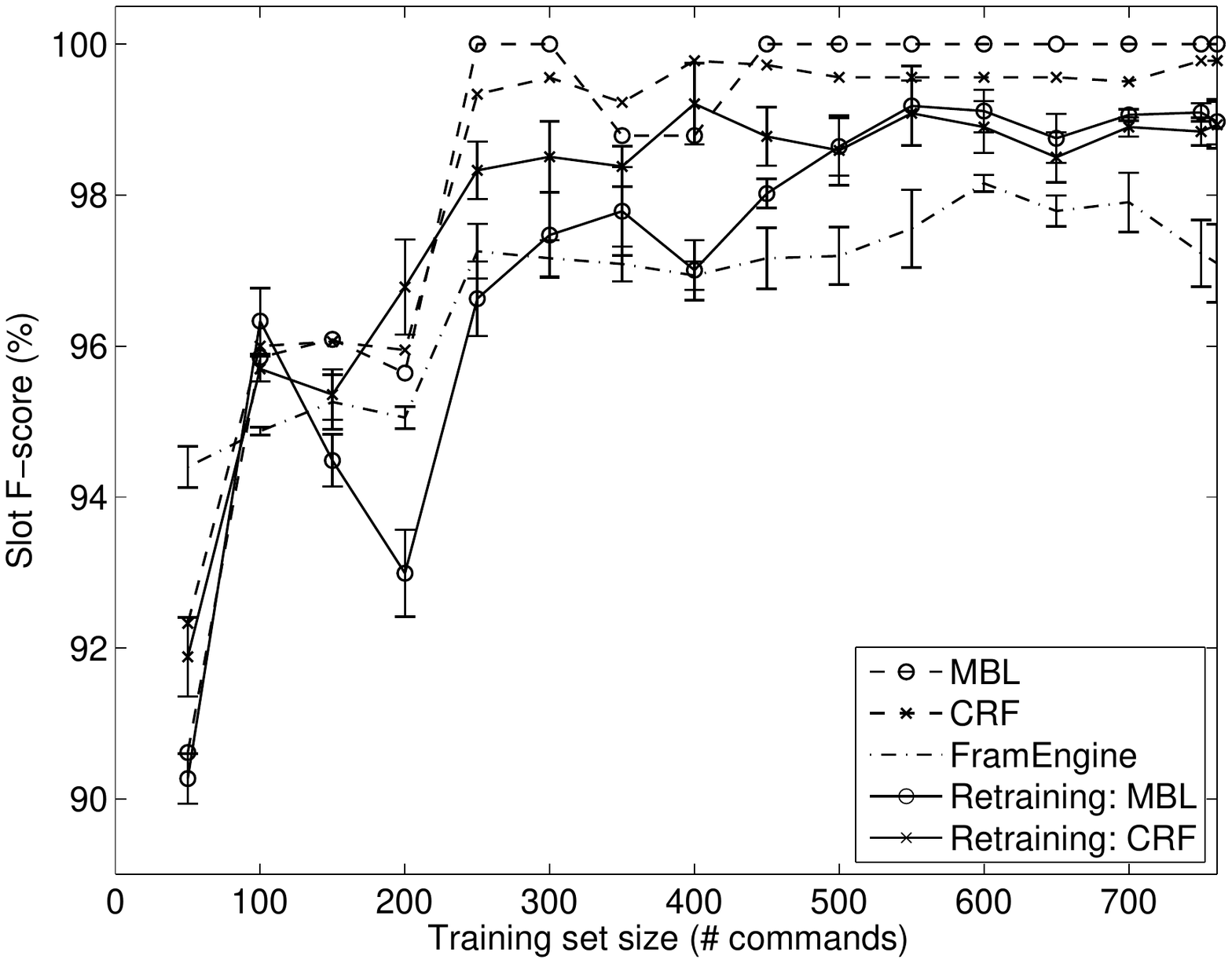} & \includegraphics[trim=50 180 50 200,clip,width=7.5cm]{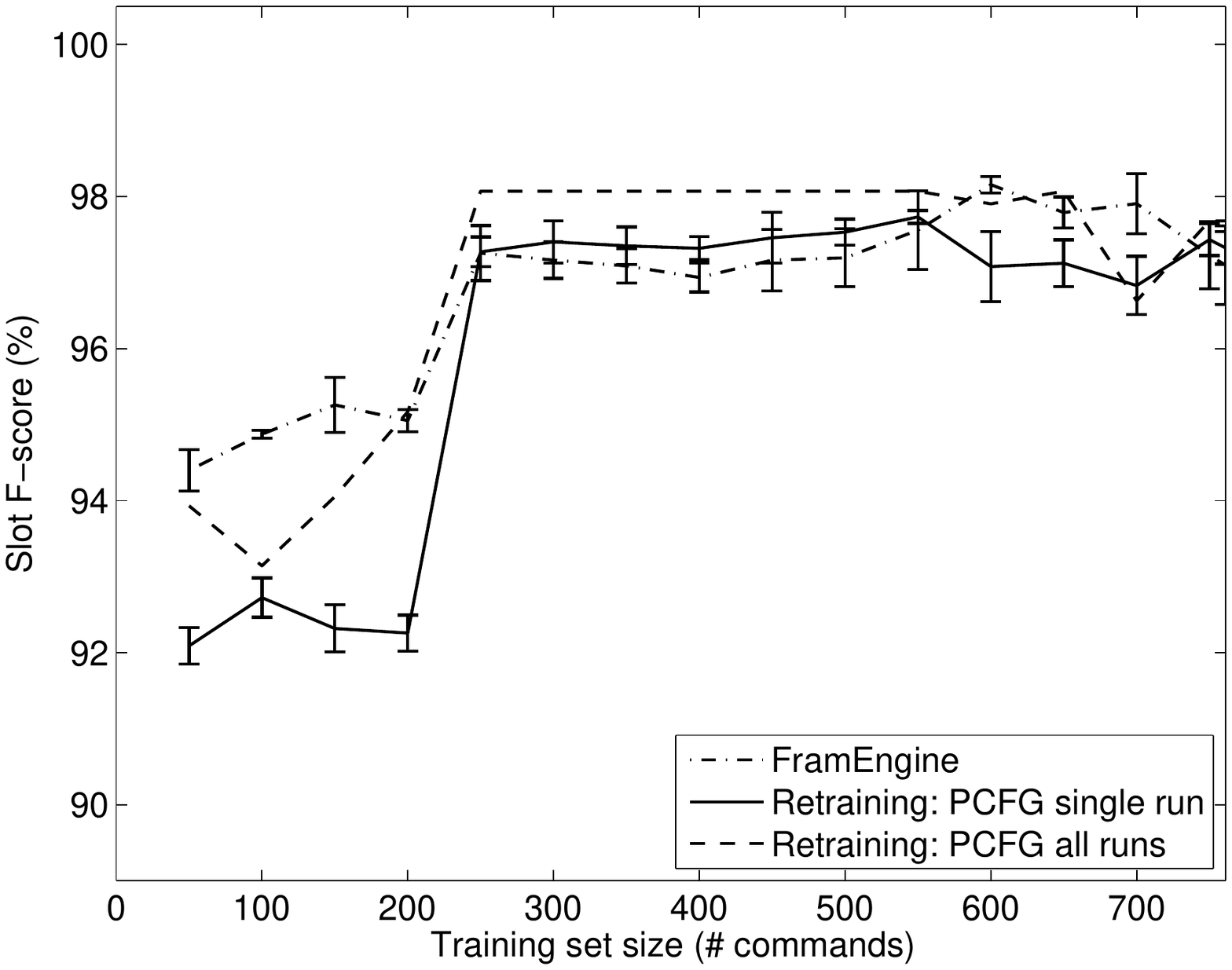}\\
(a) & (b)
\end{tabular}
\end{center}
\caption{Learning curve experiments for the original FramEngine labeling, supervised and retrained sequence labeling (a) and retraining using PCFG induction (B)}
\label{lcurves}
\end{figure*}

\subsection{Experimental Setup}

We compare four different approaches on the basis of their ability to fill slots with the correct values in semantic frames:

\begin{enumerate}
\item{Fully supervised sequence labeling ({\sc mbt} and {\sc wapiti})}
\item{FramEngine} 
\item{Retraining using sequence labeling ({\sc mbt} and {\sc wapiti})}
\item{Retraining using PCFG-induction}
\end{enumerate}

FramEngine fills frame slots by choosing slot-value tags with the highest summed probability observed during decoding, exceeding a certain threshold (established during development experiments). If no slot value tag with sufficient probability mass occurs in the utterance, the slot remains empty. Approaches 1, 3 and 4 fill frames in a similar vein, except that these techniques predict for each token in the utterance a single slot-value tag without a probability score (i.e. probability = 1). To solve ties, we add a small bias proportional to the time index to each single tag probability (t * 10\textsuperscript{-5}).

\subsection{Evaluation}

An important requirement for the ALADIN system is that the amount of effort involved for the user in training the system be kept to a minimum. To evaluate our techniques in this vein, we perform learning curve experiments to see which techniques learn the required task the fastest with the least amount of training data. The experiments were performed on the {\sc p9} set, which consists of 1142 commands and corresponding action frames of one single speaker. The last third of the data set is used as evaluation data, while the remaining training data is divided into partitions of 50 utterances. We preserve the order of the training utterances as they have been recorded, to mimic how the actual system would be trained in practice. 

The training partitions in the {\sc p9} set are used incrementally as training data for FramEngine. Optimal parameters and information sources for FramEngine were obtained during expansive experiments on the {\sc p1-8} \cite{1901.10680}. The trained hierarchical hidden markov model is then used to transcode the training utterances into slot-value tag sequences. These automatically tagged utterances, derived through a weakly supervised induction process, are then used to train the memory-based and CRF-based sequence taggers and to induce a PCFG respectively. Finally, the retrained techniques are evaluated on the held-out data, creating a data point on the learning curve.

We evaluate against gold-standard semantic frames in which only the slots mentioned in the utterance have been specified (in contrast to the overspecified frame in Figure \ref{patexample}) and which slot values can be ambiguous: e.g. in the case of 'aas opzij leggen' (`put ace aside'), the reference frame includes all four possible foundation values (target\_foundation={1,2,3,4}), because no specific value is specified in the command. We use frame slot F-score as our metric of choice, which is calculated as follows:

\begin{itemize}
\item{slot precision: \#correctly filled slots / \#total filled slots in predicted frame}
\item{slot recall: \#correctly filled slots / \#total filled slots in reference frame}
\item{slot F$_{\beta=1}$-score: 2* slot precision * slot recall /(slot precision + slot recall)}
\end{itemize}

Finally, as FramEngine's performance is affected by random initializations of the system, the experiments with FramEngine and the retrained systems are carried out ten times, with different random initializations. Scores are averaged across the ten runs and 95\% confidence levels for the scores are reported as well.

\subsection{Results}

Figure \ref{lcurves} displays the learning curves for the {\sc p9} data. We consider the fully supervised sequence labelers as the upper bound, since they are trained on gold-standard data. We notice that they achieve optimal F-score at around 250 utterances. Before that, their accuracy scores are marred by unseen words in the evaluation set, something that the weakly supervised FramEngine approach is able to overcome through its looser generalization properties. Retraining using the sequence labelers (Figure \ref{lcurves}(a)), trained on the automatically tagged data, improves F-scores for nearly all training sizes over FramEngine. For this task, retraining using conditional random fields generally yields more favorable learning curves than using memory-based learning.

The large leaps in the curves in Figure \ref{lcurves} are to a great extent related to a shift in the user's lexicon over time. In the test set (featuring recordings during the end of the data collection process), the card value `king' is always referred to as 'heer' (English: {\em Sir}), whereas initially, this card value is always referred to as 'koning' (English: `king'). The word 'heer' starts to appear in the training set after the 200th utterance, thereby causing a significant leap for all of the approaches.

Retraining through PCFG induction (Figure \ref{lcurves}(b)) similarly suffers from out-of-vocabulary words and, additionally, unseen grammatical productions in the evaluation set on smaller training sizes. Its continuing performance is fairly constant, although it seldom outperforms FramEngine in a convincing manner. Figure \ref{lcurves}(b) shows two PCFG-curves: one that expresses performance of 10 individual PCFGs for each FramEngine initialization (single run) and one that joins the data from all the runs and induces one single PCFG (all runs). Joining data over initializations dramatically improves the number of structures in the evaluation set that are covered by the grammars, while precision does not seem to suffer, thereby yielding higher F-scores for most training set sizes over 250. 

These experiments show that, although FramEngine achieves workable F-scores using a weakly supervised approach, we can automatically increase F-scores through retraining. For this task, sequence labeling is the recommended retraining technique, although PCFGs can also be observed to increase F-scores for this task. The latter result is encouraging, since it allows for the weakly supervised induction of a classic command \& control interface that restricts the acoustic search space by means of a context-free grammar.

\subsection{PCFG induction: a qualitative analysis}

\begin{figure}
\begin{center}
\small
\begin{tabular}{llll}
    {\tt ROOT} & $\rightarrow$ & {\tt DC} [0.0824324]\\
    {\tt ROOT} & $\rightarrow$ & {\tt O DC} [0.331081]\\
    {\tt ROOT} & $\rightarrow$ & {\tt F O T} [0.586486]\\
    {\tt DC} & $\rightarrow$ & {\em kaarten omdraaien} [0.593054]\\
    {\tt DC} & $\rightarrow$ & {\em nieuwe kaarten omdraaien} [0.131062]\\
    {\tt DC} & $\rightarrow$ & {\em omdraaien} [0.275885]\\
    {\tt F} & $\rightarrow$ & {\tt S V} [1.0]\\
    {\tt T} & $\rightarrow$ & {\tt TF} [0.0909515]\\
    {\tt T} & $\rightarrow$ & {\tt S V} [0.909049]\\    
    {\tt S} & $\rightarrow$ & {\tt c} [0.249707]\\
    {\tt S} & $\rightarrow$ & {\tt d} [0.264478]\\
    {\tt S} & $\rightarrow$ & {\tt h} [0.233763]\\
    {\tt S} & $\rightarrow$ & {\tt s} [0.252052]\\
    {\tt V} & $\rightarrow$ & {\tt 1} [0.0763444]\\
    {\tt V} & $\rightarrow$ & {\tt 2} [0.0617492]\\
    {\tt V} & $\rightarrow$ & {\tt 3} [0.0617492]\\
    {\tt V} & $\rightarrow$ & {\tt 4} [0.084428]\\
    {\tt V} & $\rightarrow$ & {\tt 5} [0.0911643]\\
    {\tt V} & $\rightarrow$ & {\tt 6} [0.0862243]\\
    {\tt V} & $\rightarrow$ & {\tt 7} [0.0739868]\\
    {\tt V} & $\rightarrow$ & {\tt 8} [0.0817335]\\
    {\tt V} & $\rightarrow$ & {\tt 9} [0.086112]\\ 
    {\tt V} & $\rightarrow$ & {\tt 10} [0.0793758]\\
    {\tt V} & $\rightarrow$ & {\tt 11} [0.0673627]\\
    {\tt V} & $\rightarrow$ & {\tt 12} [0.0718536]\\
    {\tt V} & $\rightarrow$ & {\tt 13} [0.0779162]\\
    {\tt O} & $\rightarrow$ & {\em kaarten} [0.13268]\\
    {\tt O} & $\rightarrow$ & {\em nieuwe} [0.252725]\\
    {\tt O} & $\rightarrow$ & {\em op} [0.614595]\\ \hline
    {\tt 1} & $\rightarrow$ & {\em aas} [1.0]\\
    {\tt 2} & $\rightarrow$ & {\em twee} [1.0]\\
    {\tt 3} & $\rightarrow$ & {\em drie} [1.0]\\
    {\tt 4} & $\rightarrow$ & {\em vier} [1.0]\\
    {\tt 5} & $\rightarrow$ & {\em vijf} [1.0]\\
    {\tt 6} & $\rightarrow$ & {\em zes} [1.0]\\
    {\tt 7} & $\rightarrow$ & {\em zeven} [1.0]\\
    {\tt 8} & $\rightarrow$ & {\em acht} [1.0]\\
    {\tt 9} & $\rightarrow$ & {\em negen} [1.0]\\
    {\tt 10} & $\rightarrow$ & {\em tien} [1.0]\\
    {\tt 11} & $\rightarrow$ & {\em boer} [1.0]\\
    {\tt 12} & $\rightarrow$ & {\em dame} [1.0]\\
    {\tt c} & $\rightarrow$ & {\em klaveren} [1.0]\\
    {\tt d} & $\rightarrow$ & {\em ruiten} [1.0]\\
    {\tt h} & $\rightarrow$ & {\em harten} [1.0]\\
    {\tt s} & $\rightarrow$ & {\em schoppen} [1.0]\\
\end{tabular}
\end{center}
\caption{Probabilistic Context-Free Grammar and lexicon (induced in a weakly supervised manner). For display purposes, only productions that occur more than 0.5\% of the time were used to compile the PCFG. The non-terminals are {\bf ROOT}, {\bf D}eal{\bf C}ard, {\bf F}rom, {\bf T}arget, {\bf S}uit, {\bf V}alue, {\bf O}. (induced in run-1 at training set size 761)}
\label{pcfgexample}
\end{figure}

While the F-scores of the PCFG-retraining approach trail compared to those of the retrained sequence labelers, the former presents an interesting opportunity for interpretability. Looking at the induced PCFGs can provide us with further insight into how the structural properties of the language are being modeled.

The grammar in Figure \ref{pcfgexample} corresponds rather well with our intuitions for this task. Apart from an artifact (production T$\rightarrow$TF cannot reach terminal nodes) and a needlessly ambiguous handling of the tokens {\em nieuwe} (new), {\em kaarten} (cards) and {\em omdraaien} (turn over), the induced PCFG presents a clean description of the grammar used in the utterances of the training set. The probabilities for the different card suits are evenly distributed, as are their card values. The two different frame types are triggered by three ROOT rules, two for the {\sc dealcard} frame, one for the {\sc movecard} frame. Interestingly, while the grammar in Figure \ref{pcfgexample} excludes infrequent rules (such as ROOT $\rightarrow$ F T O T O F O T), applying this kind of filtering in practice yields significantly lower F-scores during the experiments, as these productions provide coverage for idiosyncratic utterances containing repairs and the like.

\section{Conclusion and Future work}
\label{conclusion}

This paper described research on data collected for a command \& control interface for the card game Patience. The task of associating components of a command to slots in semantic frame representations of the intended control was already established with a fair degree of accuracy using a semantic frame induction technique based on non-negative matrix factorization and a hierarchical hidden markov model (FramEngine). This paper improves on this result by applying retraining using a discriminative sequence labeler as a post-processing step. Additional retraining experiments using probabilistic context-free grammar induction likewise show improvements when used as a post-processing step.

Next steps in this research should investigate whether these retraining techniques can yield similar accuracy increases when applied on the acoustic portion of the {\sc patcor} data. It will be particularly interesting to see whether the PCFG induction technique can yield a similarly transparent collection of productions as displayed in Figure \ref{pcfgexample}, when confronted with a noisy set of observations.

We will also investigate the advantage of the retraining techniques on other data sets, such as the grammatically less complex home automation data, collected in the context of the ALADIN project \cite{ons2013self}, but also the ROBOCUP data \cite{Chen2008}, which in recent years has become an interesting benchmark to gauge advances in this field. Finally, further research efforts should investigate whether the retraining techniques, now used as a post-processing step, can be compiled into FramEngine so that decoding can be achieved in a single processing sweep.

\section{Acknowledgments*}
This research described in this paper was funded by IWT-SBO grant 100049 (ALADIN). The {\sc patcor} dataset is available from \url{https://github.com/clips/patcor}. We would like to thank Jort Gemmeke, Hugo Van hamme and Bart Ons for their valued input on the research described in this paper.

\bibliographystyle{acl}
\bibliography{janneke}

\end{document}